\documentclass[review]{elsarticle}

\usepackage{lineno,hyperref}
\modulolinenumbers[5]

\usepackage{algorithm}
\usepackage{algpseudocode}
\usepackage{graphicx}
\usepackage{caption}
\usepackage{subcaption}
\usepackage{amsmath}
\usepackage{array}

\newcommand{\xin}[1]{\ifbool{inccomment}{{\color{blue} #1}}{}}
\newcommand{\klaus}[1]{\ifbool{inccomment}{{\color{red} #1}}{}}

\journal{Journal of Image and Vision Computing}

\begin{document}

\begin{frontmatter}

\title{Beyond saliency: understanding convolutional neural networks from saliency prediction on layer-wise relevance propagation}

\author[firstaddress]{Heyi Li\fnref{myfootnote}}
\ead{heyli@cs.stonybrook.edu}
\fntext[footnote1]{Work done during an internship at Midea Emerging Technology Center}

\author[secondaddress]{Yunke Tian}
\ead{yunke.tian@midea.com}

\author[firstaddress]{Klaus Mueller}
\ead{mueller@cs.stonybrook.edu}

\author[secondaddress]{Xin Chen\corref{mycorrespondingauthor}}
\cortext[mycorrespondingauthor]{Corresponding author}
\ead{chen1.xin@midea.com}

\address[firstaddress]{Department of Computer Science, Stony Brook University, Stony Brook, NY, USA}
\address[secondaddress]{Midea Emerging Technology Center, San Jose, CA, USA}

\begin{abstract}
Despite the tremendous achievements of deep convolutional neural networks~(CNNs) in many computer vision tasks, understanding how they actually work remains a significant challenge. In this paper, we propose a novel two-step understanding method, namely \textbf{\textit{Salient Relevance (SR) map}}, which aims to shed light on how deep CNNs recognize images and learn features from areas, referred to as \textbf{\textit{attention areas}}, therein. Our proposed method starts out with a layer-wise relevance propagation~(LRP) step which estimates a pixel-wise relevance map over the input image. Following, we construct a context-aware saliency map, SR map, from the LRP-generated map which predicts areas close to the foci of attention instead of isolated pixels that LRP reveals. In human visual system, information of regions is more important than of pixels in recognition. Consequently, our proposed approach closely simulates human recognition. Experimental results using the ILSVRC2012 validation dataset in conjunction with two well-established deep CNN models, AlexNet and VGG-16, clearly demonstrate that our proposed approach concisely identifies not only key pixels but also attention areas that contribute to the underlying neural network's comprehension of the given images. As such, our proposed SR map constitutes a convenient visual interface which unveils the visual attention of the network and reveals which type of objects the model has learned to recognize after training. The source code is available at \url{https://github.com/Hey1Li/Salient-Relevance-Propagation}.
\end{abstract}

\begin{keyword}
Convolutional Neural Networks \sep Deep Learning Understanding \sep Salient Relevance Map \sep Attention Area
\end{keyword}

\end{frontmatter}


\section{Introduction}
\label{into}
In most areas of computer vision and pattern recognition, deep convolutional neural networks (CNNs) have outperformed other methods, especially in image classification~\cite{krizhevsky2012imagenet,chen2013multi,simonyan2014very,szegedy2015going,he2016deep,tissera2016deep,guo2016deep,yu2017convolutional}. However, a conceptual understanding why these networks work so well is still largely lacking. Since visualization is an effective way to bring insight into these matters, in recent years a number of such techniques have been proposed, both from the visual analytics and the deep learning communities, aiming to understand how and why deep learning works. Visual analytics researchers typically go about the problem by designing a GUI interface with standard graphics tools to illuminate the connections among neurons intuitively \cite{harley2015interactive,liu2017towards}. Deep learning researchers, on the other hand, tend to focus on visualizing the learned features of each neuron using different optimization-based algorithms. 

The optimization-based methods are generally divided into two categories: activation maximization~\cite{erhan2009visualizing,simonyan2013deep,yosinski2015understanding} and code inversion~\cite{mahendran2016visualizing,dosovitskiy2016inverting}. In order to depict a single neuron, the strategy of activation maximization methods is to generate an image which maximally activates this specific neuron. The final result then serves as a representative of the candidate features learned by this neuron. The code inversion methods adopt a similar principle but with a different objective. At a specific layer, code inversion aims to produce an activation vector which is close to the target vector generated by a real image. In this way it is revealed how the CNNs encode the input at each layer. Unfortunately, neither of these two methods can return images recognizable to a human analyst because the solution space is so vast. This severely diminishes their usefulness. Besides, as is mentioned in \cite{wei2015understanding}, one single neuron does not necessarily respond only to one single feature but multiple features at the same time. To account for this problem, Nguyen~\textit{et al.} \cite{nguyen2016multifaceted} proposed the multifaceted feature visualization (MFV) method to visualize different facets of each neuron. Although more realistic than before, the MFV method's results are still bizarre and hallucinogenic. 

\begin{figure}
\captionsetup{width=1\textwidth}
\centering
\includegraphics[width=1\linewidth]{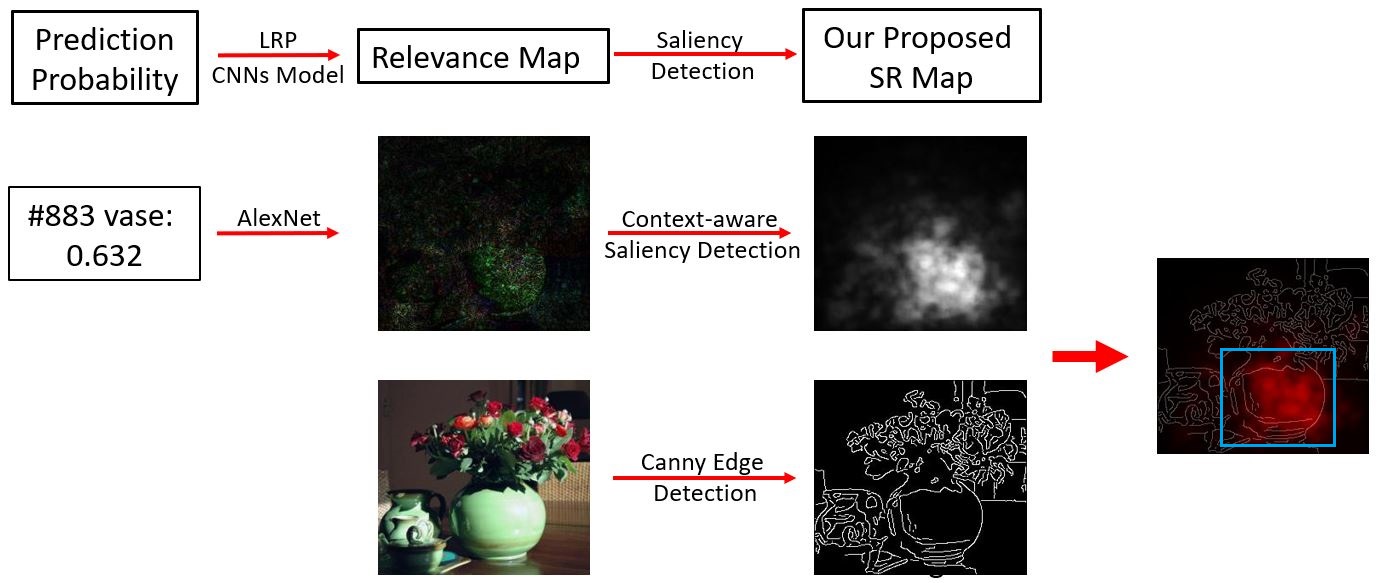}
\caption{Top row: An overview of our proposed two-step visualization scheme. Bottom row: A real example. We project our SR map onto the edge map for better visualization. The areas highlighted by the blue rectangle are attention areas. The intensity of color red is proportional to values of SR output. Best viewed electronically, in color, with zoom.}
\label{fig:1}
\end{figure}

In spite of the strong progress made in deep learning visualization \cite{kahng2018cti,pezzotti2018deepeyes}, one important aspect is largely overlooked, namely that these approaches are indirect and thus cannot specifically point out which areas of the image the network is trained to focus on. Bash~\textit{et al.} \cite{bach2015pixel} proposed the method of Layer-wise Relevance Propagation (LRP) to bridge this gap. For each input image, LRP propagates its classification probability backward through the trained network and calculates relevance scores for all pixels. By examining the intensity of different pixels in the relevance map, a direct impression of which pixels are deemed important by the network can be achieved. While this is a good step in the right direction, the LRP's relevance map fails to reveal key structural information, making the attention areas indistinguishable. This renders LRP less ideal for showing how network models perceive images (see also Section~\ref{related} for further discussion). 

To address this problem we present a new two-step approach to understand deep learning. At its core is a new map, referred to as ``Salient Relevance~(SR) Map", which directly points out the predominant attention areas of the deep network models. 
The basic scheme of our proposed method is outlined in Figure~\ref{fig:1}. First, starting out from the trained network model, we use LRP to generate a pixel relevance map for the given image subject to recognition. Second, we use a visual salience model to filter out irrelevant regions from this relevance map and thus reveal true attention areas, from which the model learns features representing the recognized object. The final result is the SR map, shown in Figure~\ref{fig:1}, second image column. 

Visual saliency is a biological framework which detects dominant foci of human attention. Simonyan et al.~\cite{simonyan2013deep} introduced saliency into deep learning visualization -- computing saliency maps for each class directly from derivatives. Due to their origin in neuroscience, saliency models accurately and reliably reflect real perception. Among saliency methods, context-aware saliency method is widely considered as one of top ones that extract salient objects together with their surroundings, making it most effective in capturing the whole area of visual attention. We therefore adopt the context-aware saliency approach~\cite{goferman2012context} to filter LRP output and reveal attention areas. 

Recently, visual attention in deep neural networks has become one of the hottest research topics. And attention mechanisms have been shown to be an essential part of the success of deep neural networks~\cite{cho2015describing,vaswani2017attention,kim2017structured} in various applications. One important property of human perception is that one does not process the whole image all at once. Instead human eyes shift attention selectively to where they is needed. To some degree, neural network models recognize objects in a similar way in that classification scores are largely determined by certain attention area rather than the whole image. This provides a strong motivation for us to adopt the context-aware saliency detection algorithm to unveil the attention areas of neural network models. 

We applied our proposed approach to several well-known CNN structures loaded with pre-trained parameters such as AlexNet and the VGG-16 on ILSVRC2012 validation dataset \cite{russakovsky2015imagenet}. Our results demonstrate that regions which belong to the recognized objects are clearly highlighted in our SR maps. This is an interesting finding because it shows that neural networks, in some sense, mimic human vision where attention mechanism and visual saliency also play an important role~\cite{itti1998model}. Finally, our results also show that our approach can clearly identify weaknesses in a given CNN model (or in the data it has been trained with). 

The major contributions of our work are twofold. 
\begin{itemize}
\item We first propose a new heatmap, Salient Relevance Map~(SR), to understand Deep Learning. The SR is generated by a two-step method which combines saliency detection with LRP. Our algorithm effectively and efficiently reveals visual attention areas of the network model, thus reflects the network's internal understanding of the input images behind its prediction. 
\item We firstly utilize ``attention" to understand and visualize CNN models. Although attention is widely used in computer vision field, to the best of our knowledge, we are the first to use attention areas to help understand and interpret how a CNN model recognizes an image. 
\end{itemize}

The remainder of the paper is structured as follows. Section~\ref{related} discusses the LRP method and models of visual saliency. Our proposed algorithm is explained thoroughly in Section~\ref{SRP}. To prove the effectiveness of our technique, we designed and performed extensive experiments which are described in Section~\ref{apps}. Section~\ref{conclusions} presents conclusions along with a brief discussion of future work. 

\section{Related Work}
\label{related}

\begin{figure}
\captionsetup{width=1\textwidth}
\centering
\includegraphics[width=.6\linewidth]{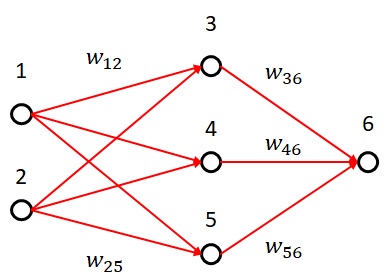}
\caption{A simple neural network with six neurons in total. }
\label{fig:2}
\end{figure}

In this section, we briefly introduce the layer-wise relevance propagation~(LRP) algorithm and review some saliency work relevant to our proposed method. 
\subsection{The Layer-Wise Relevance Propagation (LRP) algorithm}
\label{LRP}
LRP is an inverse method which calculates the contribution of a single pixel to the prediction made by the network in the image classification task. The overall idea of pixel-wise decomposition is explained in~\cite{bach2015pixel}. Here we briefly reiterate some basic concepts of LRP using a simple example. 

Given an input image $x$, a prediction score $f(x)$ is returned by the model denoted as function $f$. Suppose the network has $L$ layers, each of which is treated as a vector with dimensionality $V(l)$, where $l$ represents the index of layers. Then according to the conservation principle, LRP aims to find a relevance score $R_d$ for each vector element in layer $l$ such that the following equation holds. 

\begin{equation} \label{eq1}
f(x) = \sum_{d \in V_L} R_d^{(L)} = \cdots = \sum_{d \in V_l} R_d^{(l)} = \cdots = \sum_{d \in V_1} R_d^{(1)}
\end{equation}

As we can see in the above formula, LRP uses the prediction score as the sum of relevance scores for the last layer of the network and maintains this sum throughout all layers. Figure \ref{fig:2} shows a simple network with six neurons. $w_{ij}$ are weights and $R_i^{(l)}$ are relevance scores to be calculated. Then we have the following equation. 

\begin{equation} \label{eq2}
f(x) = R_6^{(3)} = R_3^{(2)} + R_4^{(2)} + R_5^{(2)} = R_1^{(1)} + R_2^{(1)}
\end{equation}

Furthermore, the conservation principle also guarantees that the inflow of relevance scores to one neuron equals the outflow of relevance scores from the same neuron. 

\begin{equation} \label{eq3}
R_3^{(2)} = R_{6 \rightarrow 3}^{(3) \rightarrow (2)} = R_{3 \rightarrow 1}^{(2) \rightarrow (1)} + R_{3 \rightarrow 2}^{(2) \rightarrow (1)}
\end{equation}

$R_{j \rightarrow i}^{(l+1) \rightarrow (l)}$ is the message sent from neuron $j$ at layer $l+1$ to neuron $i$ at layer $l$ and is computed using network weights according to the equation below.

\begin{equation} \label{eq4}
R_{j \rightarrow i}^{(l+1) \rightarrow (l)} = \frac{w_{ij} x_i}{\sum_k w_{kj} x_k \pm \epsilon} R_j^{(l+1)}
\end{equation}

where $\epsilon$ is to prevent numerical degenerations in case the denominator is close to zero. 

\subsection{The Saliency Map}
\label{saliencymap}
Visual saliency is a biologically inspired model of measuring which information stands out relative to its neighbors and so attracts human attention~\cite{carrasco2011visual}. It was originally promoted by psychologists in the study of attention in infancy \cite{colombo2001development}. Itti \textit{et al.}~\cite{itti1998model} presented a computational architecture to introduce the basic Koch and Ullman model \cite{koch1987shifts} to the field of computer vision. There is an extensive body of literature on various applications of visual saliency, such as~~\cite{zhu2014ensemble,oh2016detection,terzic2017texture,mukherjee2017saliency} to name just a few. Lately, researchers have also integrated saliency with the latest deep learning techniques in salient object detection\cite{han2018advanced}. Some early pioneering works directly apply CNNs to extract salient features. To better deal with saliency at multiple scale levels, \cite{li2016deep} proposes a network model with two components, one performs convolution on pixel-level and the other performs spatial pooling on patch-level. More recently, \cite{hou2017deeply} introduces skip connections among different scales and a fusion loss for combining all channels. As is summarized in~\cite{harel2007graph}, most visual saliency models are implemented in three stages: 

\begin{enumerate}[(1)]
\item \textit{extraction}: extracting low-level features over the image
\item \textit{activation}: generating activation maps from the features
\item \textit{normalization/combination}: normalizing the activation maps and combining them into a single saliency map
\end{enumerate}

Saliency detection algorithms generally place an emphasis on identifying fixation points of a human viewer and detect a single dominant object. However, apart from raw saliency, the context in which the dominant object is located is equally essential in image understanding, and this is the reason why we choose context-aware saliency detection \cite{goferman2012context} as our saliency model. The context-aware saliency detection algorithm extracts salient objects in the image together with their meaningful surroundings. As such it obeys all four psychological rules of human visual attention \cite{koffka2013principles}, making it an ideal paradigm to reveal a true visual attention. 

\begin{enumerate}[\textbf{Rule}]
\item \textbf{1:} Consider local low-level features such as contrast and color
\item \textbf{2:} Suppress frequently-occurring features
\item \textbf{3:} Organize one or several centers of visual gravity
\item \textbf{4:} Maintain high-level factors
\end{enumerate} 

Saliency methods have garnered attention among deep learning researchers \cite{sundararajan2017axiomatic,montavon2017explaining,li2017cnn,kindermans2017patternnet} because they can address all the desirabilities mentioned above. Most of this existing work tries to calculate heatmaps from the predictions of the network, albeit using different equations. 

Most recently, Kindermans \textit{et al.} pointed out that methods such as LRP may suffer from the problem of input variance~\cite{kindermans2017reliability}. We conducted our studies independently at around the same time than these authors but pursued a different and presumably more powerful variant of this approach. In our work. we go further and beyond LRP by adding saliency detection directly into the deep CNN understanding and interpretation scheme. The results we obtained using our framework show that this proposed approach is indeed highly effective. 

\begin{algorithm}
\caption{Our Proposed Two-step Algorithm to Generate SR Map}\label{alg:SR}
\begin{algorithmic}[1]
\State \textbf{Part 1: layer-wise relevance propagation}
\State Input: prediction score
\State Output: relevance map
\Procedure{calculate relevance score}{}
\For {each layer $l$}
\State $R^l = l \rightarrow lrp(R^{l-1})$
\Comment{call each layer's lrp function}
\EndFor
\EndProcedure
\State \textbf{Part 2: context-aware saliency detection}
\State Input: relevance map
\State Output: salient relevance (SR) map
\Procedure{calculate single-scale saliency}{}
\For {each pixel $i$}\Comment{$p_i$ is a patch centered at pixel $i$}
\State $d_{color}(p_i, p_j)$
\Comment{$d_{color}$ is the Euclidean distance in color space}
\State $d_{position}(p_i, p_j)$
\Comment{$d_{position}$ is the Euclidean distance of positions}
\State $d(p_i, p_j) = \frac{d_{color}(p_i, p_j)}{1 + c \cdot d_{position}(p_i, p_j)}$
\Comment{dissimilarity measure}
\EndFor
\EndProcedure
\Procedure{calculate multi-scale saliency}{}
\For {each pixel $i$}\Comment{scales at $\{r, \frac{1}{2}r, \frac{1}{4}r\}$}
\State $S_i^r = 1 - exp\{-\frac{1}{K} \sum_{k=1}^{K} d(p_i^r, q_k^{r_k})\}$
\Comment{K most similar patches}
\State $\bar{S_i} = \frac{1}{M} \sum\limits_{r \in R} S_i^r$
\Comment{mean saliency at different scales}
\EndFor
\EndProcedure
\Procedure{include immediate context}{}
\For {each pixel $i$}\Comment{$\bar{S_i} > 0.8$}
\State $S_i = \bar{S_i}(1-d_{foci}(i))$
\Comment{$d_{foci}$ is the Euclidean distance between pixel $i$ and closest focus of attention pixel}
\EndFor
\EndProcedure
\end{algorithmic}
\end{algorithm}

\section{Methodology}
\label{SRP}
\subsection{Algorithm for Generating the Salient Relevance (SR) Map}
\label{description}
We shall now describe the algorithm that generates our proposed Salient Relevance (SR) map. Figure \ref{fig:1} provides an illustration of our workflow, using an example image of a vase with flowers. Algorithm~\ref{alg:SR} lists the pseudo code.  

In the first step we generate the standard relevance map using the layer-wise relevance propagation (LRP) algorithm. The input is the classification vector calculated by the network model from the input image (the bottom left image in Figure \ref{fig:1}). To precisely infer the model's perception of the input image, only the class of the highest probability value is retained while other values are set to zero. This probability is then propagated backward through the network, layer-by-layer. At each layer, the perception information is transferred from the network's output feature maps to the input feature maps using the existing parameters. The propagation continues until the first layer of the model has been reached. The relevance map generated in this way is shown in the top left of Figure \ref{fig:1}. It has the same size as the original image. 

The second step refines the relevance map into our \textit{salient} relevance (SR) map. We achieve this using context-aware saliency detection, as explained next.
In context-aware saliency detection a single pixel is considered salient if the patch centered at this pixel is distinct from other image patches, and is so at multiple scales.
It is also important to note that background patches tend to remain similar at different scales while objects in the foreground are likely to be salient only at a few scales. This helps to distinguish background from foreground regions. In order to include background regions surrounding the foci of attention, each pixel outside the attended areas is weighted based on its Euclidean distance to the closest attended pixel. In this way, interesting background areas of salient objects are incorporated into the saliency map, while non-interesting regions are excluded. Via this procedure, the method then uses the information gathered at multiple scales to increase the contrast between salient and non-salient pixels. The salient relevance map for our vase with flowers example is shown in the center column, top image of Figure \ref{fig:1},    

Finally, we integrate the salient relevance (SR) map with a Canny edge map obtained from the input image to afford a more contextual visualization of the SR map. Figure \ref{fig:1}, center column, bottom image shows the Canny edge map, while Figure \ref{fig:1}, right image, shows the Canny edge map fused with the SR map. 

\subsection{Comparative Experiments}
\label{comparison}
\begin{figure}
\captionsetup{width=1\textwidth}
\centering

\begin{subfigure}{.4\textwidth}
\centering
\includegraphics[width=.9\linewidth]{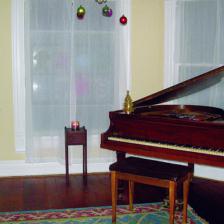}
\caption{original image}
\label{fig3:sub1}
\end{subfigure}%
\begin{subfigure}{.4\textwidth}
\includegraphics[width=.9\linewidth]{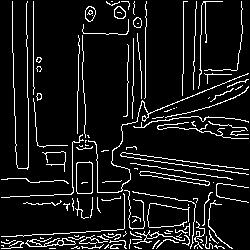}
\centering
\caption{edge map}
\label{fig3:sub2}
\end{subfigure}

\begin{subfigure}{.4\textwidth}
\includegraphics[width=.9\linewidth]{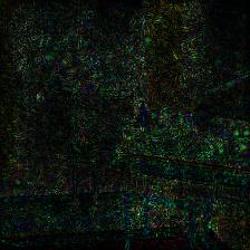}
\centering
\caption{LRP relevance map}
\label{fig3:sub3}
\end{subfigure}%
\begin{subfigure}{.4\textwidth}
\includegraphics[width=.9\linewidth]{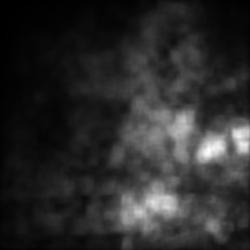}
\centering
\caption{Our proposed SR map}
\label{fig3:sub4}
\end{subfigure}

\begin{subfigure}{.4\textwidth}
\centering
\includegraphics[width=.9\linewidth]{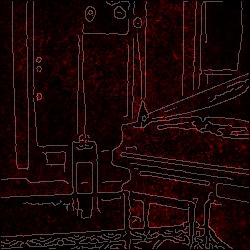}
\caption{LRP relevance map with edge}
\label{fig3:sub5}
\end{subfigure}%
\begin{subfigure}{.4\textwidth}
\includegraphics[width=.9\linewidth]{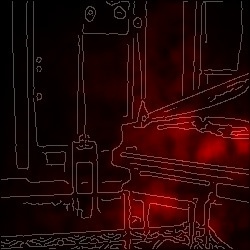}
\centering
\caption{SR map with edges}
\label{fig3:sub6}
\end{subfigure}

\caption{The ground truth label of the original image is "grand piano". We first apply layer-wise relevance propagation algorithm to produce the LRP relevance map shown in \ref{fig3:sub3}. Then we perform saliency detection on the relevance map to produce our proposed SR map in \ref{fig3:sub4}. Finally, we project both LRP relevance map and SR map onto the edge map for better visual comparison. From \ref{fig3:sub5} and \ref{fig3:sub6}, it is obvious that the SR map is superior at uncovering the network model's true understanding. Best viewed electronically, in color.}
\label{fig:3}
\end{figure}

In the following we use two running examples to illustrate our proposed method, and also compare it with the conventional one. The examples will show that our method better unveils the model's real perception. 
The first study uses the pre-trained AlexNet \cite{krizhevsky2012imagenet} as the network model and images from ILSVRC2012 validation dataset as the input. Results are shown in Figure \ref{fig:3}. 


This image is labeled "grand piano" as ground truth, and AlexNet classifies it correctly. Using LRP, we can propagate the prediction probability backward through the network to see which pixels contribute to the classification result. It is fair to assume that those pixels should fall within the area of the piano in the image since the network recognizes this object. However, as we can see in \ref{fig3:sub3}, it is not the case. Although pixels which represent the piano are present in the LRP relevance map, many irrelevant pixels are also included such as those on top. In fact, those unrelated pixels are so prominent that it is quite difficult to distinguish the piano from the relevance map.

To determine the exact attention area, we apply the saliency detection procedure on the relevance map.
This generates the SR map where we observe that all of the dominant pixels in fact belong to the detected object. The SR map clearly shows the attention area of the network model which is the piano in the front. 
Next, we fuse both the relevance map and the SR map with the Canny edge map \cite{canny1987computational} of the original image, with the aim to better visualize the maps in the context of the scene's boundaries. Comparing \ref{fig3:sub5} and \ref{fig3:sub6} it is obvious that our method outperforms LRP at precisely revealing the attention area of the network. 

\begin{figure}
\captionsetup{width=1\textwidth}
\centering

\begin{subfigure}{.4\textwidth}
\centering
\includegraphics[width=.9\linewidth]{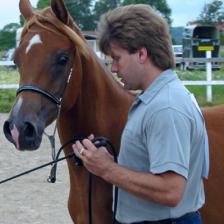}
\caption{original image}
\label{fig4:sub1}
\end{subfigure}%
\begin{subfigure}{.4\textwidth}
\includegraphics[width=.9\linewidth]{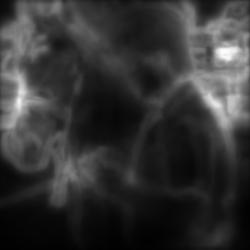}
\centering
\caption{saliency map}
\label{fig4:sub2}
\end{subfigure}

\begin{subfigure}{.4\textwidth}
\includegraphics[width=.9\linewidth]{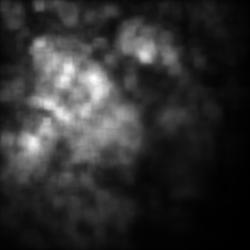}
\centering
\caption{Our proposed SR map}
\label{fig4:sub3}
\end{subfigure}%
\begin{subfigure}{.4\textwidth}
\includegraphics[width=.9\linewidth]{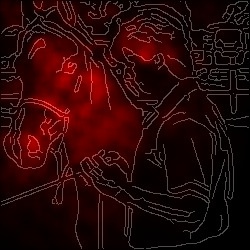}
\centering
\caption{SR map with edges}
\label{fig4:sub4}
\end{subfigure}

\caption{We first apply saliency detection on the original image to generate the saliency map shown in Figure \ref{fig4:sub2}. The saliency map shows that the horse, the man and the horse trailer are most salient to the human eyes. Then we 
use LRP to produce the relevance map which is not shown here for conciseness. The SR map is derived from the relevance map by performing saliency detection. Figure \ref{fig4:sub4} is formed by projecting the proposed SR map onto the Canny edge map of the original image. By comparing Figure \ref{fig4:sub2} and Figure \ref{fig4:sub4}, we clearly show that the human eyes and the network model focus on different objects. Best viewed electronically, in color.}
\label{fig:4}
\end{figure}

In order to show that our method successfully captures the network's perception focus of an image, we compare the saliency map of the original image with our SR map. The former represents what human eyes would recognize while the latter shows what the network detects. Figure \ref{fig:4} shows the difference. 


As we can see in Figure \ref{fig4:sub2}, the dominant objects in the attention area
include the horse, the man, and the horse trailer further away in the back. Since among these three dominant objects only the horse  is included in the ImageNet training dataset, it comes at no surprise  that our pre-trained AlexNet model classifies the image as ``sorrel" which is a breed of horse named after its hair coat color. As we can see in our SR map (which is derived from the relevance map), only pixels which belong to the horse are prominent. Neither the man nor the horse trailer are present. One thing to note is that the man's hair is also highlighted in our SR map. That is because the man's hair shares a similar color and texture as the horse's hair. These results vividly prove that our SR map is capable of effectively showing the network's real understanding. 

\begin{figure}
\captionsetup{width=1\textwidth}
\centering

\begin{subfigure}{.3\textwidth}
\centering
\includegraphics[width=.9\linewidth]{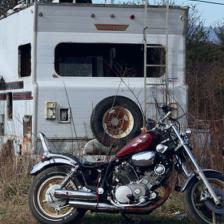}
\caption{GT: RV}
\label{fig5:sub11}
\end{subfigure}%
\begin{subfigure}{.3\textwidth}
\centering
\includegraphics[width=.9\linewidth]{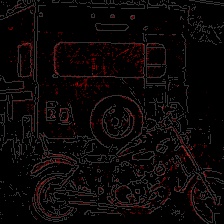}
\caption{Deep SHAP}
\label{fig5:sub12}
\end{subfigure}%
\begin{subfigure}{.3\textwidth}
\centering
\includegraphics[width=.9\linewidth]{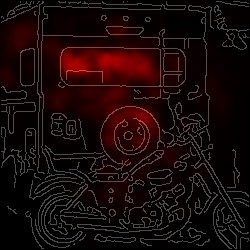}
\caption{SR map}
\label{fig5:sub13}
\end{subfigure}

\begin{subfigure}{.3\textwidth}
\centering
\includegraphics[width=.9\linewidth]{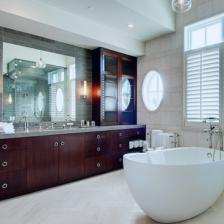}
\caption{GT: bathtub}
\label{fig5:sub21}
\end{subfigure}%
\begin{subfigure}{.3\textwidth}
\centering
\includegraphics[width=.9\linewidth]{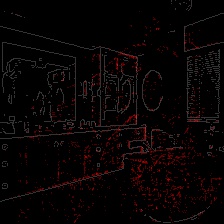}
\caption{Deep SHAP}
\label{fig5:sub22}
\end{subfigure}%
\begin{subfigure}{.3\textwidth}
\centering
\includegraphics[width=.9\linewidth]{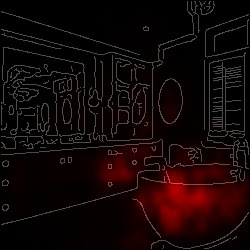}
\caption{SR map}
\label{fig5:sub23}
\end{subfigure}

\begin{subfigure}{.3\textwidth}
\centering
\includegraphics[width=.9\linewidth]{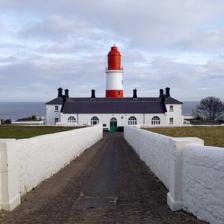}
\caption{GT: lighthouse}
\label{fig5:sub31}
\end{subfigure}%
\begin{subfigure}{.3\textwidth}
\centering
\includegraphics[width=.9\linewidth]{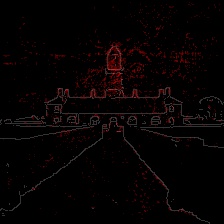}
\caption{Deep SHAP}
\label{fig5:sub32}
\end{subfigure}%
\begin{subfigure}{.3\textwidth}
\centering
\includegraphics[width=.9\linewidth]{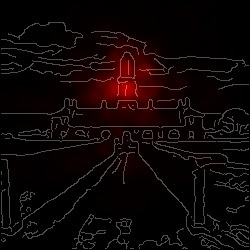}
\caption{SR map}
\label{fig5:sub33}
\end{subfigure}

\begin{subfigure}{.3\textwidth}
\centering
\includegraphics[width=.9\linewidth]{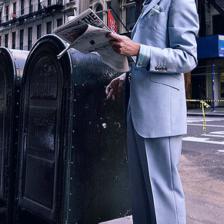}
\caption{GT: mailbox}
\label{fig5:sub41}
\end{subfigure}%
\begin{subfigure}{.3\textwidth}
\centering
\includegraphics[width=.9\linewidth]{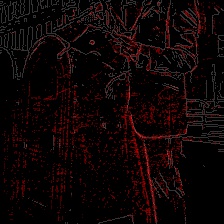}
\caption{Deep SHAP}
\label{fig5:sub42}
\end{subfigure}%
\begin{subfigure}{.3\textwidth}
\centering
\includegraphics[width=.9\linewidth]{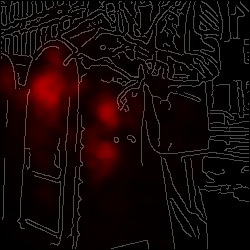}
\caption{SR map}
\label{fig5:sub43}
\end{subfigure}

\caption{The first column shows the input image and its ground truth (GT). The second column and the third column show the corresponding results with edges using Deep SHAP and our proposed SR map respectively. In all cases, VGG16 correctly classifies the image. By comparing the second and the third column, we clearly prove that our method outperforms the Deep SHAP in explaining the hidden mechanism behind VGG16's success. Best viewed electronically, in color, with zoom.}
\label{fig:5}
\end{figure}

We choose the Deep SHAP algorithm \cite{lundberg2017unified} as another benchmark for comparison \addtocounter{footnote}{-1}\footnote{\url{https://github.com/slundberg/shap/}}. In this experiment, the network model is the pre-trained VGG16 \cite{simonyan2014very} and the input are images from the ILSVRC2012 validation dataset. Results are shown in Figure \ref{fig:5}. In the first example, the ground truth label for the original image is RV. Since VGG16 makes the correct prediction, the attention area should be able to explain its success. However, as we can see in the second column, Deep SHAP includes irrelevant pixels belonging to the motorcycle in front of the RV. In contrary, our proposed SR map only highlights regions of the RV, which best explains VGG16's hidden mechanism. Although Deep SHAP unifies six existing feature attribution methods including LRP, it does not incorporate any visual saliency model. Therefore, our algorithm outperforms Deep SHAP in better revealing network models' true foci of attention. Three other examples are shown in Figure \ref{fig:5}. 

\begin{figure}
\captionsetup{width=1\textwidth}
\centering

\begin{subfigure}{.25\textwidth}
\centering
\includegraphics[width=.9\linewidth]{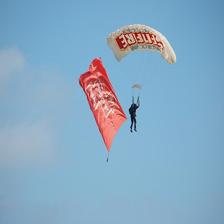}
\caption{parachute}
\label{fig6:sub11}
\end{subfigure}%
\begin{subfigure}{.25\textwidth}
\centering
\includegraphics[width=.9\linewidth]{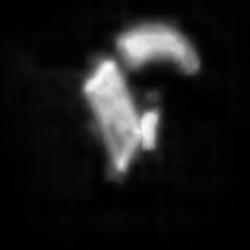}
\caption{saliency map}
\label{fig6:sub12}
\end{subfigure}%
\begin{subfigure}{.25\textwidth}
\centering
\includegraphics[width=.9\linewidth]{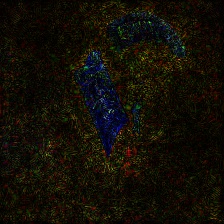}
\caption{LRP}
\label{fig6:sub13}
\end{subfigure}%
\begin{subfigure}{.25\textwidth}
\centering
\includegraphics[width=.9\linewidth]{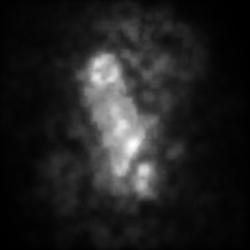}
\caption{SR map}
\label{fig6:sub14}
\end{subfigure}

\begin{subfigure}{.25\textwidth}
\centering
\includegraphics[width=.9\linewidth]{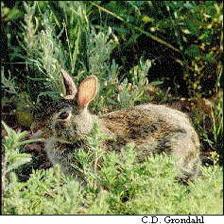}
\caption{wood rabbit}
\label{fig6:sub21}
\end{subfigure}%
\begin{subfigure}{.25\textwidth}
\centering
\includegraphics[width=.9\linewidth]{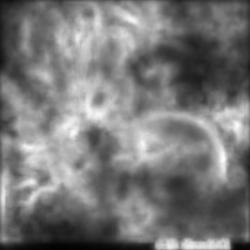}
\caption{saliency map}
\label{fig6:sub22}
\end{subfigure}%
\begin{subfigure}{.25\textwidth}
\centering
\includegraphics[width=.9\linewidth]{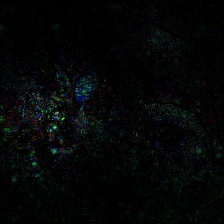}
\caption{LRP}
\label{fig6:sub23}
\end{subfigure}%
\begin{subfigure}{.25\textwidth}
\centering
\includegraphics[width=.9\linewidth]{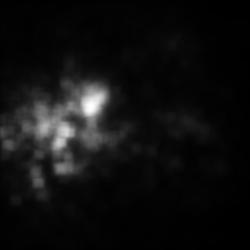}
\caption{SR map}
\label{fig6:sub24}
\end{subfigure}

\caption{In the first example, there is one parachuter in the blue sky. The saliency map is simple and clear. Our proposed SR map reveals that the network focuses on the left parachute. The SSIM value between our proposed SR map and the saliency map is high. In the second example, a rabbit is half hidden among bushes. This complex input image causes the saliency map to be messy. Our proposed SR map shows that the network captures the head and the ears of the rabbit without background bushes. The SSIM value between our proposed SR map and the saliency map is low.}
\label{fig:6}
\end{figure}

For quantitative comparisons, we choose the structure similarity index (SSIM) \cite{wang2004image} as a metric. We apply the context-aware saliency detection algorithm directly on the original input images and use these saliency maps as the gold standard because saliency algorithms convey the true visual perception of human observers. Then we calculate the SSIM value of the LRP relevance map and our SR map with the reference saliency map, respectively. Higher SSIM values indicate better perception and thus are more preferable. We use the pre-trained AlexNet as the model and randomly select $500$ images from ILSVRC2012 validation dataset as the input. Based on our experiment results, we find that the SSIM value varies significantly given different input images. Two examples are shown in Figure \ref{fig:6}. This is because when the input image has complex background, the original saliency map contains both foreground features and background features. The network, on the other hand, is trained to only capture the main object. As a result, there is a gap between the saliency map and our proposed SR map, which leads to a low SSIM value. Therefore, instead of comparing the absolute SSIM values directly, we choose the ratio of these two SSIM values. As is shown in Table \ref{table1}, our SR map consistently outscores the LRP relevance map. 

\begin{table}
\begin{center}
\begin{tabular}{ |m{6em} | m{3cm} | m{3cm} | m{2cm} | } 
\hline
 & $SSIM_{1}$ (LRP and saliency map) & $SSIM_{2}$ (SR and saliency map) & $\frac{SSIM_{2}}{SSIM_{1}}$ ratio\\ 
\hline\hline
parachute & 0.2756 & \textbf{0.5040} & \textbf{1.8287} \\
\hline
wood rabbit & 0.0490 & \textbf{0.0828} & \textbf{1.6898} \\ 
\hline
average ratio of $500$ images & N/A & N/A & \textbf{1.7038} \\
\hline
\end{tabular}
\end{center}
\caption{The first two rows correspond to the two examples in Figure \ref{fig:6}. Given different input images, the SSIM values vary significantly. However, our SR map consistently achieves higher SSIM values, compared to the LRP relevance map. We randomly select $500$ images from the ILSVRC2012 validation dataset and compute the average ratio of two SSIM values for each input image. As is shown above, our proposed SR map reliably outperforms the LRP relevance map.}
\label{table1}
\end{table}

These examples support the claim that the CNN model treats visual input in a fashion quite similar to the human visual system. The latter has been well studied. In that sense, our SR map, derived from the LRP-generated relevance map, provides the link among these two systems, the CNN and the real network that is part of the human visual system. 

\section{Case Studies and Discussions}
\label{apps}
In this section. we further explore potential applications of the SR map via several case studies. All of our experiments were conducted using a NVIDIA GTX 1080 graphics card on Ubuntu 16.04 LTS. We implemented the layer-wise relevance propagation (LRP) algorithm \addtocounter{footnote}{-1}\footnote{\url{http://www.heatmapping.org/tutorial/}} using the PyTorch \footnote{\url{https://github.com/pytorch/pytorch}} package. We used the Matlab code \footnote{Code available at \url{http://webee.technion.ac.il/cgm/Computer-Graphics-Multimedia/Software/Saliency/Saliency.html}} offered by the original authors of the context-aware saliency detection scheme. Our neural networks are pre-trained models included in the PyTorch package whose benchmark performances on ILSVRC 2012 validation dataset are also publicly available \footnote{\url{https://github.com/jcjohnson/cnn-benchmarks}}. Our code is written in Jupyter Notebook and can be obtained from Github \footnote{\url{https://github.com/Hey1Li/Salient-Relevance-Propagation}}, which includes the settings for all our experiments. Since we use pre-trained models for our experiments, we do not need to train the neural networks from scratch. The inference time is also negligible because it is comparable to one backpropagation step without updating any parameters. The saliency detection step, on the other hand, takes up most of the testing time. In our case, the running time of one image using context-aware saliency detection algorithm varies from $40$ seconds to $200$ seconds depending on image complexity.

\begin{figure}
\captionsetup{width=1\textwidth}
\centering

\begin{subfigure}{.4\textwidth}
\centering
\includegraphics[width=.9\linewidth]{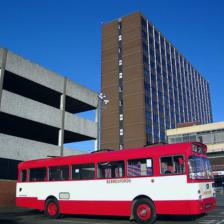}
\caption{original image}
\label{fig7:sub1}
\end{subfigure}%
\begin{subfigure}{.4\textwidth}
\includegraphics[width=.9\linewidth]{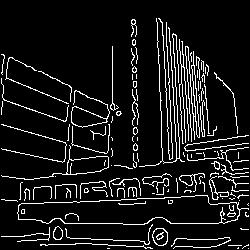}
\centering
\caption{edge map}
\label{fig7:sub2}
\end{subfigure}

\begin{subfigure}{.4\textwidth}
\centering
\includegraphics[width=.9\linewidth]{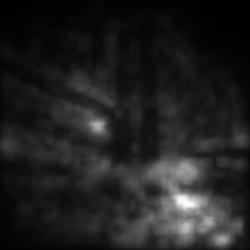}
\caption{SR map of AlexNet}
\label{fig7:sub3}
\end{subfigure}%
\begin{subfigure}{.4\textwidth}
\centering
\includegraphics[width=.9\linewidth]{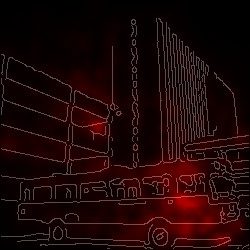}
\caption{SR map of AlexNet with edges}
\label{fig7:sub4}
\end{subfigure}

\begin{subfigure}{.4\textwidth}
\centering
\includegraphics[width=.9\linewidth]{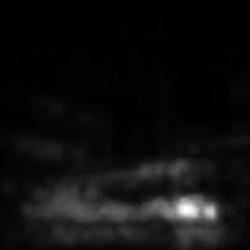}
\caption{SR map of VGG-16}
\label{fig7:sub5}
\end{subfigure}%
\begin{subfigure}{.4\textwidth}
\centering
\includegraphics[width=.9\linewidth]{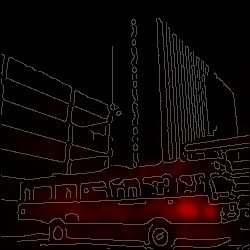}
\caption{SR map of VGG-16 with edges}
\label{fig7:sub6}
\end{subfigure}

\caption{In the input image, there is a red bus with a white stripe parked in front of some buildings. As is shown in the SR map, AlexNet fails to separate the bus from the buildings. Thus it gives the wrong prediction ``cinema". On the contrary, VGG-16 has no trouble to distinguish the two objects. Only the bus part is highlighted in the SR map. As a result, VGG-16 successfully labels it as ``coach". Best viewed electronically, in color.}
\label{fig:7}
\end{figure} 

\begin{figure}
\captionsetup{width=1\textwidth}
\centering

\begin{subfigure}{.3\textwidth}
\centering
\includegraphics[width=.9\linewidth]{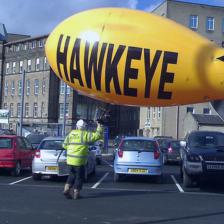}
\caption{GT: airship}
\label{fig8:sub11}
\end{subfigure}%
\begin{subfigure}{.3\textwidth}
\centering
\includegraphics[width=.9\linewidth]{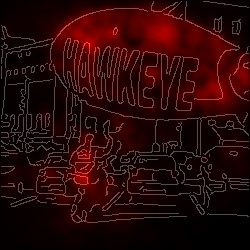}
\caption{Alex: school bus}
\label{fig8:sub12}
\end{subfigure}%
\begin{subfigure}{.3\textwidth}
\centering
\includegraphics[width=.9\linewidth]{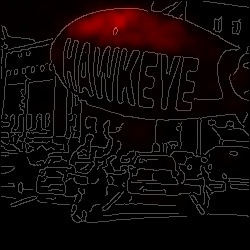}
\caption{VGG: airship}
\label{fig8:sub13}
\end{subfigure}

\begin{subfigure}{.3\textwidth}
\centering
\includegraphics[width=.9\linewidth]{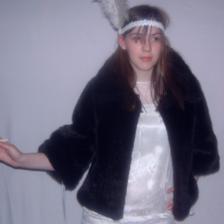}
\caption{GT: fur coat}
\label{fig8:sub21}
\end{subfigure}%
\begin{subfigure}{.3\textwidth}
\centering
\includegraphics[width=.9\linewidth]{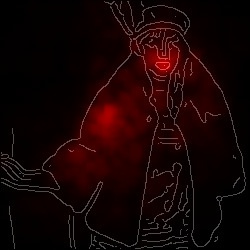}
\caption{Alex: chimpanzee}
\label{fig8:sub22}
\end{subfigure}%
\begin{subfigure}{.3\textwidth}
\centering
\includegraphics[width=.9\linewidth]{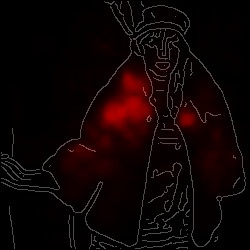}
\caption{VGG: fur coat}
\label{fig8:sub23}
\end{subfigure}

\begin{subfigure}{.3\textwidth}
\centering
\includegraphics[width=.9\linewidth]{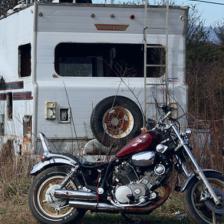}
\caption{GT: RV}
\label{fig8:sub31}
\end{subfigure}%
\begin{subfigure}{.3\textwidth}
\centering
\includegraphics[width=.9\linewidth]{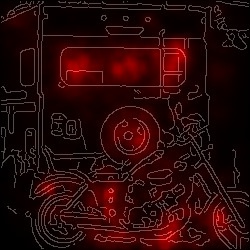}
\caption{Alex: Model T}
\label{fig8:sub32}
\end{subfigure}%
\begin{subfigure}{.3\textwidth}
\centering
\includegraphics[width=.9\linewidth]{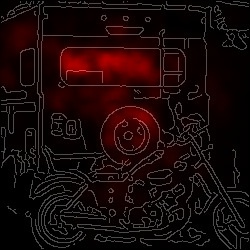}
\caption{VGG: RV}
\label{fig8:sub33}
\end{subfigure}

\begin{subfigure}{.3\textwidth}
\centering
\includegraphics[width=.9\linewidth]{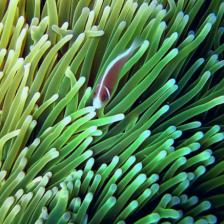}
\caption{GT: anemone fish}
\label{fig8:sub41}
\end{subfigure}%
\begin{subfigure}{.3\textwidth}
\centering
\includegraphics[width=.9\linewidth]{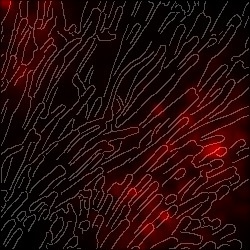}
\caption{Alex: sea anemone}
\label{fig8:sub42}
\end{subfigure}%
\begin{subfigure}{.3\textwidth}
\centering
\includegraphics[width=.9\linewidth]{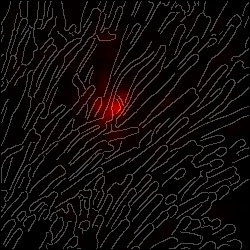}
\caption{VGG: anemone fish}
\label{fig8:sub43}
\end{subfigure}

\caption{The first column shows the input image and its ground truth (GT). The second column and the third column show the corresponding SR map of AlexNet and VGG-16 with their predictions respectively. VGG-16 gives right answers to all four images while AlexNet is wrong. In all cases, AlexNet fails to separate different objects: the yellow airship and the human in the first image, the black fur coat the girl's face in the second image, the RV and the motorcycle in the third image, the fish and the anemone in the fourth image. Best viewed electronically, in color, with zoom.}
\label{fig:8}
\end{figure}

\begin{figure}
\captionsetup{width=1\textwidth}
\centering

\begin{subfigure}{.3\textwidth}
\centering
\includegraphics[width=.9\linewidth]{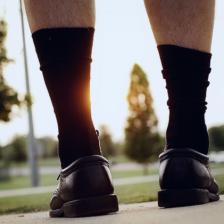}
\caption{GT: sock}
\label{fig9:sub11}
\end{subfigure}%
\begin{subfigure}{.3\textwidth}
\centering
\includegraphics[width=.9\linewidth]{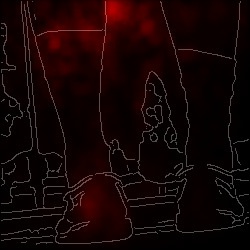}
\caption{Alex: cowboy boot}
\label{fig9:sub12}
\end{subfigure}%
\begin{subfigure}{.3\textwidth}
\centering
\includegraphics[width=.9\linewidth]{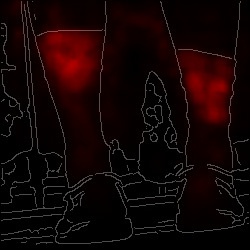}
\caption{VGG: sock}
\label{fig9:sub13}
\end{subfigure}

\begin{subfigure}{.3\textwidth}
\centering
\includegraphics[width=.9\linewidth]{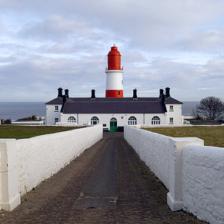}
\caption{GT: lighthouse}
\label{fig9:sub21}
\end{subfigure}%
\begin{subfigure}{.3\textwidth}
\centering
\includegraphics[width=.9\linewidth]{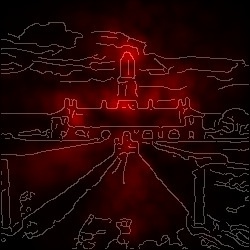}
\caption{Alex: seawall}
\label{fig9:sub22}
\end{subfigure}%
\begin{subfigure}{.3\textwidth}
\centering
\includegraphics[width=.9\linewidth]{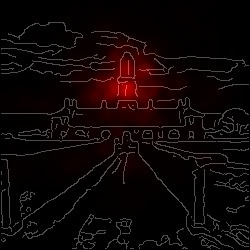}
\caption{VGG: lighthouse}
\label{fig9:sub23}
\end{subfigure}

\begin{subfigure}{.3\textwidth}
\centering
\includegraphics[width=.9\linewidth]{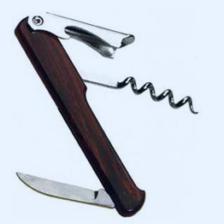}
\caption{GT: corkscrew}
\label{fig9:sub31}
\end{subfigure}%
\begin{subfigure}{.3\textwidth}
\centering
\includegraphics[width=.9\linewidth]{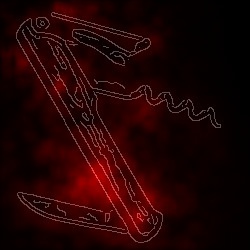}
\caption{Alex: hammer}
\label{fig9:sub32}
\end{subfigure}%
\begin{subfigure}{.3\textwidth}
\centering
\includegraphics[width=.9\linewidth]{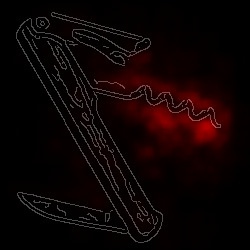}
\caption{VGG: corkscrew}
\label{fig9:sub33}
\end{subfigure}

\begin{subfigure}{.3\textwidth}
\centering
\includegraphics[width=.9\linewidth]{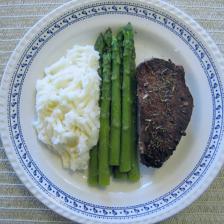}
\caption{GT: mashed potato}
\label{fig9:sub41}
\end{subfigure}%
\begin{subfigure}{.3\textwidth}
\centering
\includegraphics[width=.9\linewidth]{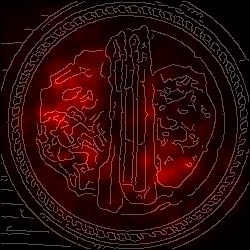}
\caption{Alex: plate}
\label{fig9:sub42}
\end{subfigure}%
\begin{subfigure}{.3\textwidth}
\centering
\includegraphics[width=.9\linewidth]{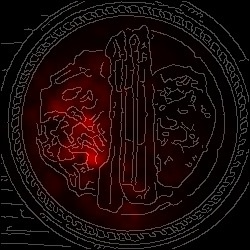}
\caption{VGG: mashed potato}
\label{fig9:sub43}
\end{subfigure}

\caption{The first column shows the input image and its ground truth (GT). The second column and the third column show the corresponding SR map of AlexNet and VGG-16 with their predictions respectively. VGG-16 gives right answers to all four images while AlexNet is wrong. In all cases, AlexNet fails to separate different objects: the shoes, the socks and the leg in the first image, the wall, the house and the lighthouse in the second image, the handle and the tools in the third image, different dishes in the fourth image. Best viewed electronically, in color, with zoom.}
\label{fig:9}
\end{figure}

\subsection{AlexNet vs. VGG-16}

In the first experiment, we analyze the performance gap between AlexNet \cite{krizhevsky2012imagenet} and VGG-16 \cite{simonyan2014very}. According to the benchmark, AlexNet achieves $42.9\%$ top-1 error rate while VGG-16's top-1 error rate is only $27\%$. Therefore, in order to explain why VGG-16 outperforms AlexNet on ImageNet, we apply our algorithm on images mislabeled by AlexNet but correctly classified by VGG-16 and examine their SR maps. The example shown in Figure \ref{fig:7} is comprehensively reviewed here. More images are included in Figure \ref{fig:8} and Figure \ref{fig:9}.

The input image is labeled ``coach" as its ground truth. VGG-16 recognizes it correctly, but AlexNet misinterpreted it as ``cinema". At first glance, it is difficult to understand why AlexNet can make such a mistake because a cinema is hugely different from a transit bus. But by looking at the SR map of AlexNet, it is obvious that some prominent pixels belong to the bus while others belong to the buildings in the back. Therefore, it is reasonable to infer that AlexNet actually identifies the coach as part of the whole structure. Since it treats the red bus as the marquee above the entrance, AlexNet thus classifies the image as a cinema with high confidence. On the contrary, VGG-16 successfully separates the bus from the building because all dominant pixels in SR map fall within the area of the coach. As we can see in Figure \ref{fig:8} and Figure \ref{fig:9}, this is not an isolated incident. VGG-16 persistently outperforms AlexNet in dividing different objects. This is why VGG-16 beats AlexNet in this test. 

\subsection{Evaluation of VGG-16}

\begin{figure}
\captionsetup{width=1\textwidth}
\centering

\begin{subfigure}{.4\textwidth}
\centering
\includegraphics[width=.9\linewidth]{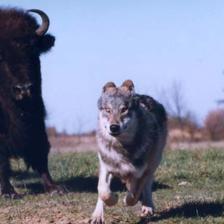}
\caption{original image}
\label{fig10:sub1}
\end{subfigure}%
\begin{subfigure}{.4\textwidth}
\includegraphics[width=.9\linewidth]{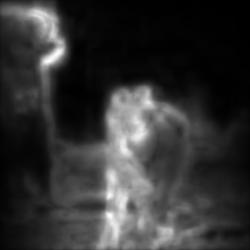}
\centering
\caption{saliency map}
\label{fig10:sub2}
\end{subfigure}

\begin{subfigure}{.4\textwidth}
\centering
\includegraphics[width=.9\linewidth]{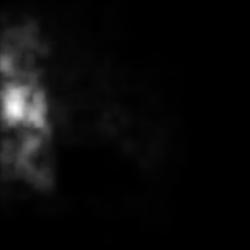}
\caption{SR map of VGG-16}
\label{fig10:sub3}
\end{subfigure}%
\begin{subfigure}{.4\textwidth}
\centering
\includegraphics[width=.9\linewidth]{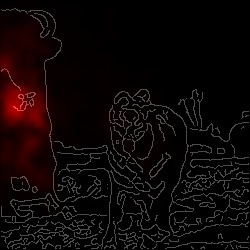}
\caption{SR map with edges}
\label{fig10:sub4}
\end{subfigure}

\caption{In this input image, there is a gray wolf in the center. It is the most prominent object to human eyes as is shown in the saliency map. We can also see part of a black ox which seems to be chasing the wolf. Our SR map reveals that VGG-16 prioritizes the bull instead of the wolf. Therefore it classifies the image as ``ox" while the correct label is ``gray wolf". In the last image, we project the SR map onto the edge map and display it in teh red channel for better visualization. Best viewed electronically, in color.}
\label{fig:10}
\end{figure}

\begin{figure}
\captionsetup{width=1\textwidth}
\centering

\begin{subfigure}{.3\textwidth}
\centering
\includegraphics[width=.9\linewidth]{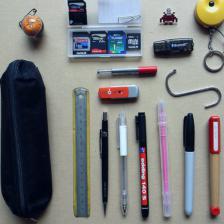}
\caption{GT: ruler}
\label{fig11:sub11}
\end{subfigure}%
\begin{subfigure}{.3\textwidth}
\centering
\includegraphics[width=.9\linewidth]{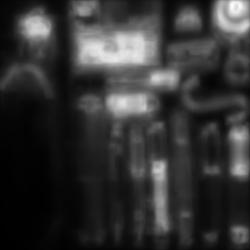}
\caption{saliency map}
\label{fig11:sub12}
\end{subfigure}%
\begin{subfigure}{.3\textwidth}
\centering
\includegraphics[width=.9\linewidth]{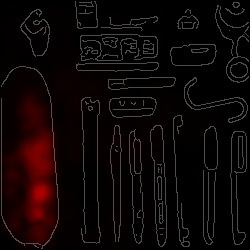}
\caption{VGG: pencil box}
\label{fig11:sub13}
\end{subfigure}

\begin{subfigure}{.3\textwidth}
\centering
\includegraphics[width=.9\linewidth]{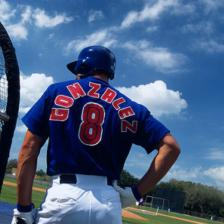}
\caption{GT: baseball player}
\label{fig11:sub21}
\end{subfigure}%
\begin{subfigure}{.3\textwidth}
\centering
\includegraphics[width=.9\linewidth]{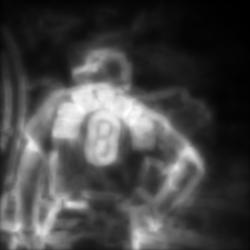}
\caption{saliency map}
\label{fig11:sub22}
\end{subfigure}%
\begin{subfigure}{.3\textwidth}
\centering
\includegraphics[width=.9\linewidth]{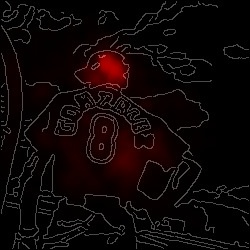}
\caption{VGG: football helmet}
\label{fig11:sub23}
\end{subfigure}

\begin{subfigure}{.3\textwidth}
\centering
\includegraphics[width=.9\linewidth]{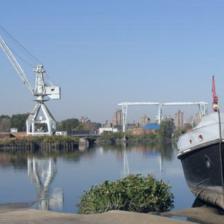}
\caption{GT: crane}
\label{fig11:sub31}
\end{subfigure}%
\begin{subfigure}{.3\textwidth}
\centering
\includegraphics[width=.9\linewidth]{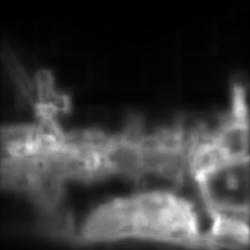}
\caption{saliency map}
\label{fig11:sub32}
\end{subfigure}%
\begin{subfigure}{.3\textwidth}
\centering
\includegraphics[width=.9\linewidth]{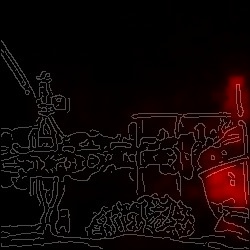}
\caption{VGG: dock}
\label{fig11:sub33}
\end{subfigure}

\begin{subfigure}{.3\textwidth}
\centering
\includegraphics[width=.9\linewidth]{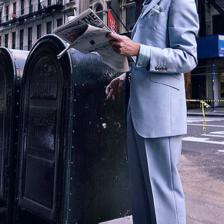}
\caption{GT: suit}
\label{fig11:sub41}
\end{subfigure}%
\begin{subfigure}{.3\textwidth}
\centering
\includegraphics[width=.9\linewidth]{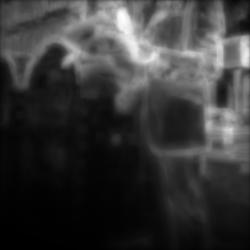}
\caption{saliency map}
\label{fig11:sub42}
\end{subfigure}%
\begin{subfigure}{.3\textwidth}
\centering
\includegraphics[width=.9\linewidth]{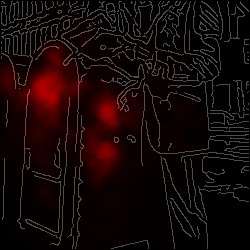}
\caption{VGG: mailbox}
\label{fig11:sub43}
\end{subfigure}

\caption{The first column shows the input image and its ground truth (GT). The second column and the third column show the saliency map and SR map of VGG-16 with its prediction respectively. VGG-16 gives wrong predictions in all four cases. Comparing it with human eyes, VGG-16 only detects a single object without proper context: the pencil box in the first image, the helmet in the second image, the ship in the third image, the mailbox in the fourth image. Best viewed electronically, in color, with zoom.}
\label{fig:11}
\end{figure}

\begin{figure}
\captionsetup{width=1\textwidth}
\centering

\begin{subfigure}{.3\textwidth}
\centering
\includegraphics[width=.9\linewidth]{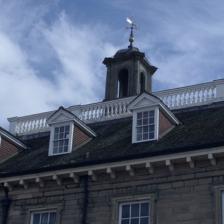}
\caption{GT: tile roof}
\label{fig12:sub11}
\end{subfigure}%
\begin{subfigure}{.3\textwidth}
\centering
\includegraphics[width=.9\linewidth]{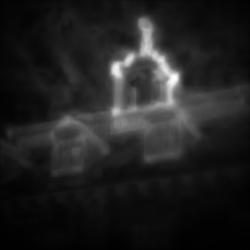}
\caption{saliency map}
\label{fig12:sub12}
\end{subfigure}%
\begin{subfigure}{.3\textwidth}
\centering
\includegraphics[width=.9\linewidth]{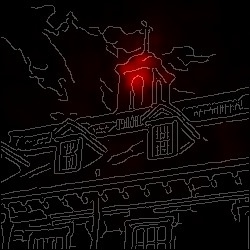}
\caption{VGG: monastery}
\label{fig12:sub13}
\end{subfigure}

\begin{subfigure}{.3\textwidth}
\centering
\includegraphics[width=.9\linewidth]{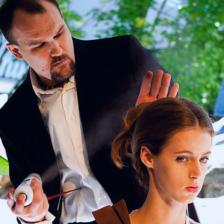}
\caption{GT: hair spray}
\label{fig12:sub21}
\end{subfigure}%
\begin{subfigure}{.3\textwidth}
\centering
\includegraphics[width=.9\linewidth]{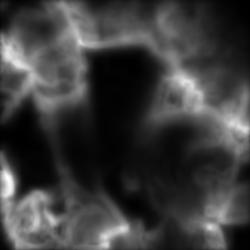}
\caption{saliency map}
\label{fig12:sub22}
\end{subfigure}%
\begin{subfigure}{.3\textwidth}
\centering
\includegraphics[width=.9\linewidth]{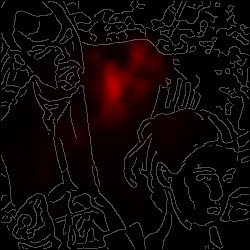}
\caption{VGG: suit}
\label{fig12:sub23}
\end{subfigure}

\begin{subfigure}{.3\textwidth}
\centering
\includegraphics[width=.9\linewidth]{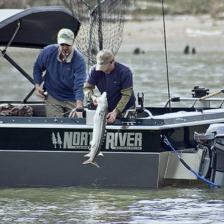}
\caption{GT: sturgeon}
\label{fig12:sub31}
\end{subfigure}%
\begin{subfigure}{.3\textwidth}
\centering
\includegraphics[width=.9\linewidth]{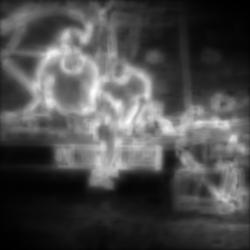}
\caption{saliency map}
\label{fig12:sub32}
\end{subfigure}%
\begin{subfigure}{.3\textwidth}
\centering
\includegraphics[width=.9\linewidth]{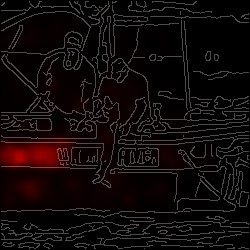}
\caption{VGG: trimaran}
\label{fig12:sub33}
\end{subfigure}

\begin{subfigure}{.3\textwidth}
\centering
\includegraphics[width=.9\linewidth]{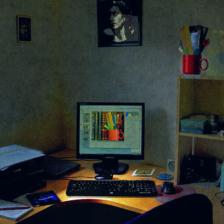}
\caption{GT: monitor}
\label{fig12:sub41}
\end{subfigure}%
\begin{subfigure}{.3\textwidth}
\centering
\includegraphics[width=.9\linewidth]{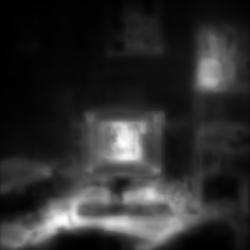}
\caption{saliency map}
\label{fig12:sub42}
\end{subfigure}%
\begin{subfigure}{.3\textwidth}
\centering
\includegraphics[width=.9\linewidth]{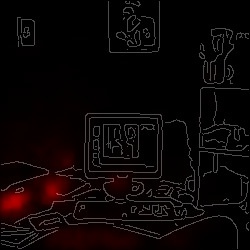}
\caption{VGG: desk}
\label{fig12:sub43}
\end{subfigure}

\caption{The first column shows the input image and its ground truth (GT). The second column and the third column show the saliency map and SR map of VGG-16 with its prediction respectively. VGG-16 gives wrong predictions in all four cases. Comparing it with human eyes, VGG-16 only detects a single object without proper context: the cupola in the first image, the suit in the second image, the trimaran in the third image, the desk in the fourth image. Best viewed electronically, in color, with zoom.}
\label{fig:12}
\end{figure}

We have explained how VGG-16 makes correct decisions in the previous experiment. But VGG-16 still mislabels about one fourth of all the validation images. As a result, we utilize the SR map to analyze VGG-16's mistakes in this study. The input image in Figure \ref{fig:10} depicts a vivid scene where a gray wolf is running away from an ox. The wolf is located in the center and is most attractive to human attention indicated by the saliency map. However, VGG-16 turns its attention to the black ox on the left side since only that area is highlighted in the SR map. This coincides with the fact that VGG-16 labels this image as ``ox" while the ground truth is ``gray wolf". So unlike AlexNet which fails to differentiate multiple objects, VGG-16 is only capable of recognizing single object from the input image and it ignores the proper context information. This is the reason why VGG-16 is wrong in this case. Since ground truth labels are based on human consensus, we claim that VGG-16 fails to grab true meanings of input images despite its capability of recognizing objects. More examples are included in Figure \ref{fig:11} and Figure \ref{fig:12}. 

\subsection{VGG-16 vs. VGG-Face}

\begin{figure}
\captionsetup{width=1\textwidth}
\centering

\begin{subfigure}{.4\textwidth}
\centering
\includegraphics[width=.9\linewidth]{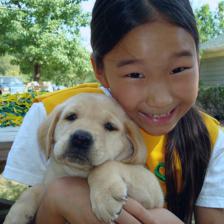}
\caption{original image}
\label{fig13:sub1}
\end{subfigure}%
\begin{subfigure}{.4\textwidth}
\includegraphics[width=.9\linewidth]{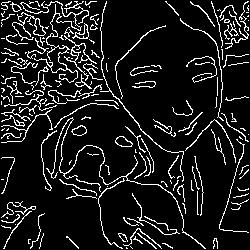}
\centering
\caption{edge map}
\label{fig13:sub2}
\end{subfigure}

\begin{subfigure}{.4\textwidth}
\centering
\includegraphics[width=.9\linewidth]{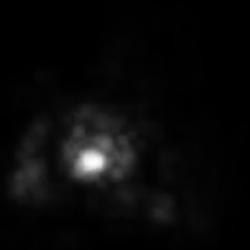}
\caption{SR map of VGG-16}
\label{fig13:sub3}
\end{subfigure}%
\begin{subfigure}{.4\textwidth}
\centering
\includegraphics[width=.9\linewidth]{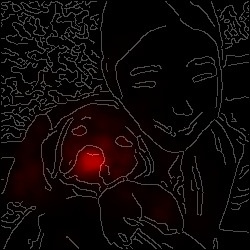}
\caption{SR map of VGG-16 with edges}
\label{fig13:sub4}
\end{subfigure}

\begin{subfigure}{.4\textwidth}
\centering
\includegraphics[width=.9\linewidth]{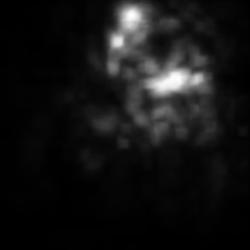}
\caption{SR map of VGG-Face}
\label{fig13:sub5}
\end{subfigure}%
\begin{subfigure}{.4\textwidth}
\centering
\includegraphics[width=.9\linewidth]{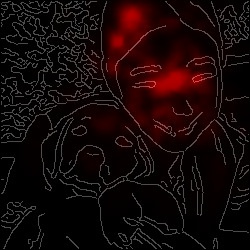}
\caption{SR map of VGG-F with edges}
\label{fig13:sub6}
\end{subfigure}

\caption{There is a teenage girl holding a cute puppy in the input image. VGG-16 is trained on ImageNet and no human faces are included in the training dataset. As a result, VGG-16 does not respond to the girl's face even though it is visually obvious. On the other hand, VGG-Face only has exposure to human faces in training. So it completely ignores the dog and only highlights the girl's face. Best viewed electronically, in color.}
\label{fig:13}
\end{figure}

\begin{figure}
\captionsetup{width=1\textwidth}
\centering

\begin{subfigure}{.3\textwidth}
\centering
\includegraphics[width=.9\linewidth]{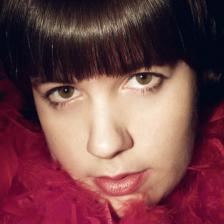}
\caption{original image}
\label{fig14:sub11}
\end{subfigure}%
\begin{subfigure}{.3\textwidth}
\centering
\includegraphics[width=.9\linewidth]{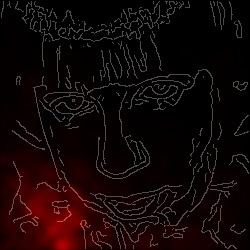}
\caption{VGG-16}
\label{fig14:sub12}
\end{subfigure}%
\begin{subfigure}{.3\textwidth}
\centering
\includegraphics[width=.9\linewidth]{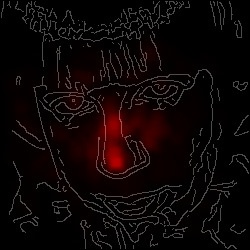}
\caption{VGG-Face}
\label{fig14:sub13}
\end{subfigure}

\begin{subfigure}{.3\textwidth}
\centering
\includegraphics[width=.9\linewidth]{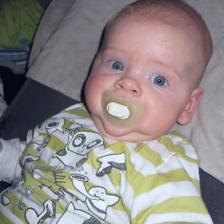}
\caption{original image}
\label{fig14:sub21}
\end{subfigure}%
\begin{subfigure}{.3\textwidth}
\centering
\includegraphics[width=.9\linewidth]{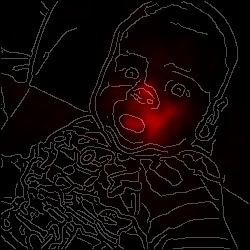}
\caption{VGG-16}
\label{fig14:sub22}
\end{subfigure}%
\begin{subfigure}{.3\textwidth}
\centering
\includegraphics[width=.9\linewidth]{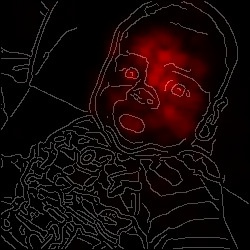}
\caption{VGG-Face}
\label{fig14:sub23}
\end{subfigure}

\begin{subfigure}{.3\textwidth}
\centering
\includegraphics[width=.9\linewidth]{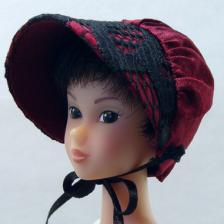}
\caption{original image}
\label{fig14:sub31}
\end{subfigure}%
\begin{subfigure}{.3\textwidth}
\centering
\includegraphics[width=.9\linewidth]{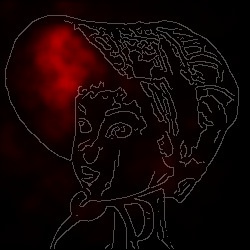}
\caption{VGG-16}
\label{fig14:sub32}
\end{subfigure}%
\begin{subfigure}{.3\textwidth}
\centering
\includegraphics[width=.9\linewidth]{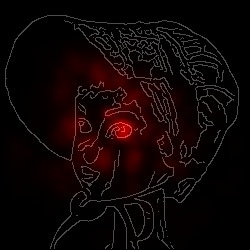}
\caption{VGG-Face}
\label{fig14:sub33}
\end{subfigure}

\begin{subfigure}{.3\textwidth}
\centering
\includegraphics[width=.9\linewidth]{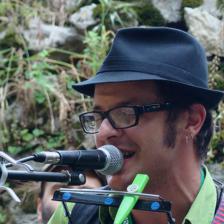}
\caption{original image}
\label{fig14:sub41}
\end{subfigure}%
\begin{subfigure}{.3\textwidth}
\centering
\includegraphics[width=.9\linewidth]{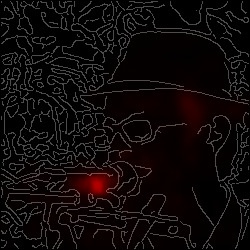}
\caption{VGG-16}
\label{fig14:sub42}
\end{subfigure}%
\begin{subfigure}{.3\textwidth}
\centering
\includegraphics[width=.9\linewidth]{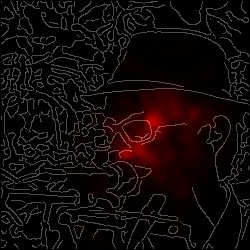}
\caption{VGG-Face}
\label{fig14:sub43}
\end{subfigure}

\caption{The first column shows the input image. The second column and the third column show the corresponding SR map of VGG-16 and VGG-Face respectively. In all four cases, VGG-Face successfully recognizes the human face while VGG-16 captures a different object: the feather boa in the first image, the pacifier in the second image, the bonnet in the third image, the microphone in the fourth image. Best viewed electronically, in color, with zoom.}
\label{fig:14}
\end{figure}

\begin{figure}
\captionsetup{width=1\textwidth}
\centering

\begin{subfigure}{.3\textwidth}
\centering
\includegraphics[width=.9\linewidth]{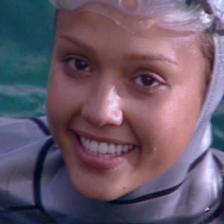}
\caption{original image}
\label{fig15:sub11}
\end{subfigure}%
\begin{subfigure}{.3\textwidth}
\centering
\includegraphics[width=.9\linewidth]{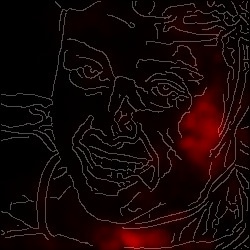}
\caption{VGG-16}
\label{fig15:sub12}
\end{subfigure}%
\begin{subfigure}{.3\textwidth}
\centering
\includegraphics[width=.9\linewidth]{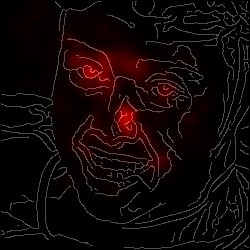}
\caption{VGG-Face}
\label{fig15:sub13}
\end{subfigure}

\begin{subfigure}{.3\textwidth}
\centering
\includegraphics[width=.9\linewidth]{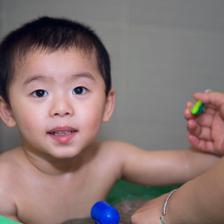}
\caption{original image}
\label{fig15:sub21}
\end{subfigure}%
\begin{subfigure}{.3\textwidth}
\centering
\includegraphics[width=.9\linewidth]{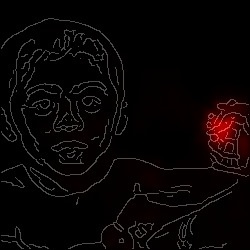}
\caption{VGG-16}
\label{fig15:sub22}
\end{subfigure}%
\begin{subfigure}{.3\textwidth}
\centering
\includegraphics[width=.9\linewidth]{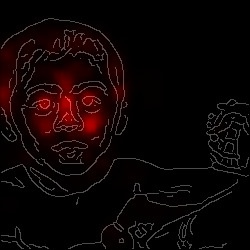}
\caption{VGG-Face}
\label{fig15:sub23}
\end{subfigure}

\begin{subfigure}{.3\textwidth}
\centering
\includegraphics[width=.9\linewidth]{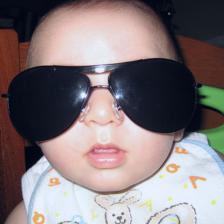}
\caption{original image}
\label{fig15:sub31}
\end{subfigure}%
\begin{subfigure}{.3\textwidth}
\centering
\includegraphics[width=.9\linewidth]{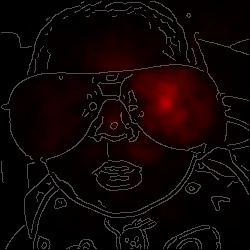}
\caption{VGG-16}
\label{fig15:sub32}
\end{subfigure}%
\begin{subfigure}{.3\textwidth}
\centering
\includegraphics[width=.9\linewidth]{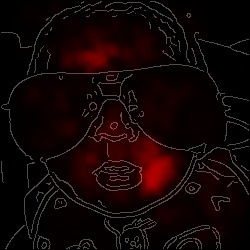}
\caption{VGG-Face}
\label{fig15:sub33}
\end{subfigure}

\begin{subfigure}{.3\textwidth}
\centering
\includegraphics[width=.9\linewidth]{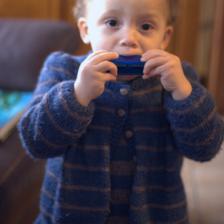}
\caption{original image}
\label{fig15:sub41}
\end{subfigure}%
\begin{subfigure}{.3\textwidth}
\centering
\includegraphics[width=.9\linewidth]{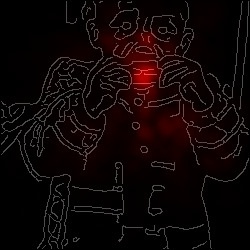}
\caption{VGG-16}
\label{fig15:sub42}
\end{subfigure}%
\begin{subfigure}{.3\textwidth}
\centering
\includegraphics[width=.9\linewidth]{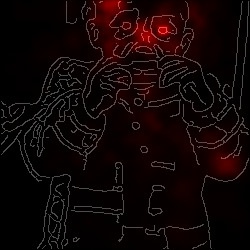}
\caption{VGG-Face}
\label{fig15:sub43}
\end{subfigure}

\caption{The first column shows the input image. The second column and the third column show the corresponding SR map of VGG-16 and VGG-Face respectively. In all four cases, VGG-Face successfully recognizes the human face while VGG-16 captures a different object: the swimming gear in the first image, the toy in the second image, the sunglasses in the third image, the harp in the fourth image. Best viewed electronically, in color, with zoom.}
\label{fig:15}
\end{figure}

The goal of our final experiment is to use the SR map to demonstrate the versatility of neural network models. VGG-16 and VGG-Face \cite{parkhi2015deep} share the same 16-layers network structure. The difference is that VGG-Face is trained using a dataset consisted of $2.6$ million facial images while VGG-16 is trained using ImageNet with no exposure to human faces. Thus given the same input image with a girl holding a puppy, these two network models are expected to have different attention areas and highlight various regions. As is shown in Figure \ref{fig:13}, our method visibly shows this distinction. VGG-Face only emphasizes the girl's face in SR map. VGG-16, on the other hand, focuses on the dog's face and label it as ``Labrador Retriever" which is one of the most popular types of dog in North America. Figure \ref{fig:14} and Figure \ref{fig:15} include more examples where all the input images contain one human face and another prominent object. 

To summarize, we conducted three experiments to demonstrate our method's capability of visualizing network models' real comprehension. In the first experiment, we compared AlexNet with VGG-16 and explained their performance gap on the ImageNet classification task. We then further analyzed VGG-16 by examining its mistakes. In the last study, we show that the same network structure can correspond to varied objects if trained on different datasets. And our experiment results evidently prove the effectiveness of our algorithm.

\section{Conclusion}
\label{conclusions}
In this paper, we have successfully proposed a novel two-step visualization algorithm to generate SR map \addtocounter{footnote}{-5}\footnote{source code available at \url{https://github.com/Hey1Li/Salient-Relevance-Propagation}}, which aims to understand deep CNN models and reveal areas from which the models learn representative features. These areas are referred to as attention areas. By combining layer-wise relevance propagation with context-aware saliency detection, our proposed method successfully reveals a CNN model's visual attention and thus true perception of input images. Experimental results using several well-known models on the ILSVRC2012 validation dataset have shown that SR map not only is capable of revealing neural network's perception but also is a superior tool for helping researchers understand deep learning models. 

In the future, we plan to apply our method to analyze performances of more complex neural network models such as ResNet. Further, we will build direct connections between our visual analysis and proper training adjustments. As a consequence, our visualization tool can be directly applied to improve performances of deep network models. 

\section*{Acknowledgements}

This work is supported by Midea Corporate Research Center University Program. We would also like to show our gratitude to Prof. Tengyu Ma for his comments on an earlier version of the manuscript. 

This research was also partially supported by NSF grant IIS 1527200, as well as the MSIP (Ministry of Science, ICT and Future Planning), Korea, under the ``ITCCP Program" directed by NIPA.

\section*{References}

\bibliography{mybibfile}

\end{document}